\documentclass[11pt]{article}
\usepackage[final]{acl}
\usepackage{times}
\usepackage{latexsym}
\usepackage[T1]{fontenc}
\usepackage[utf8]{inputenc}
\usepackage{microtype}
\usepackage{inconsolata}
\usepackage{graphicx}
\usepackage{booktabs}
\usepackage{array}
\usepackage{amsmath}
\usepackage{url}

\title{Where Does the Asymmetry Enter? Staging a Fairness Audit Across a Fused Multimodal Moderation Pipeline}

\author{Kardelen Subasi \and Jaros{\l}aw A. Chudziak \\
  Institute of Computer Science \\
  Faculty of Electronics and Information Technology \\
  Warsaw University of Technology \\
  Warsaw, Poland \\
  \texttt{\{kardelen.subasi.stud, jaroslaw.chudziak\}@pw.edu.pl}}

\begin{document}
\maketitle

\begin{abstract}
Almost every short-video upload now passes through a classifier that fuses video, speech, and on-screen text into one decision about whether it may be seen. It moderates every language community on the platform at once, and fairness across them is judged, if at all, from one accuracy score.
That score is compatible with error falling almost entirely on one language group. An audit that goes looking reads the model's decisions, the last stage of a long pipeline and the one whose test has the fewest items.
We ask where in a fused pipeline a subgroup asymmetry can be detected at all: in what the classifier may read, in what its branches encode, or in what it decides.
We answer this with a post-hoc audit of a published TikTok classifier, holding all three stages to one significance standard using only its logged decisions, labels, and pre-fusion features, with source language as the subgroup axis.
The asymmetry is significant at the stage deciding what the model may read, and again in what the text branch encodes. At the earlier stage, a language-blind cap admits less of a Vietnamese clip than an English one, for reasons of segmentation, not verbosity. The asymmetry is undetectable at the decision, where the smallest subgroup leaves twelve items to test.
An audit reading only decisions looks where its evidence is thinnest; the same test staged upstream recovers a disparity the decision-level view returns as noise.
\end{abstract}

\section{Introduction}

Whether an automated moderation system treats different groups of users fairly has become a first-order question for the short-video platforms that now police uploads with multimodal classifiers, fusing video frames, audio, and on-screen or spoken text into a single flagging decision \citep{nguyen2025mtikguard, das2023hatemm}. One such classifier is the gate through which speech in dozens of languages reaches an audience, applying a single learned notion of harm to all of it, on platforms where hateful content reaches large and young audiences and where automating the work carries its own governance hazards \citep{weimann2020tiktok, gonzalez2023populist, gorwa2020algorithmic}. These systems are evaluated, reported, and deployed on the strength of a single aggregate accuracy score, typically in the high 80s or low 90s.

\begin{figure}[!ht]
\centering
\includegraphics[width=\columnwidth]{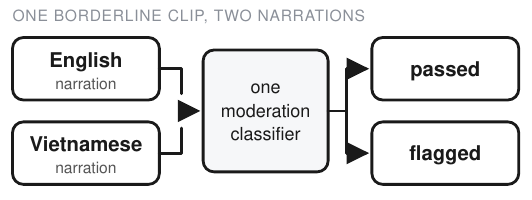}
\caption{The case the audit is built to surface. Two creators upload near-identical borderline clips, one narrated in English and one in Vietnamese, to the same classifier and receive opposite outcomes. One aggregate accuracy score is reported whichever way the errors fall, whether they are spread evenly across the language groups or concentrated almost entirely on one, so only disaggregation separates the two situations.}
\label{fig:motivation}
\end{figure}

That figure is compatible with many different underlying error distributions. A classifier reporting 90\% accuracy overall can make almost all of its errors against a few subgroups while remaining, from the outside, indistinguishable from one that errs uniformly. Consider two creators uploading near-identical borderline clips, the same mild-profanity gaming rant narrated once in Vietnamese and once in English, to the same classifier. If the Vietnamese clip is flagged while the English one is quietly passed, the two face different moderation odds despite comparable content, and the difference is invisible to the aggregate report unless someone disaggregates by language. This case is what the audit is built to surface (Figure~\ref{fig:motivation}), and we return to it throughout.

Fairness auditing outside moderation has shown such concentration to be common \citep{buolamwini2018gendershades, mehrabi2021survey}, and a few studies find the same inside multimodal systems built for other tasks \citep{jiang2023evaluating, booth2021bias} and inside text-only moderation APIs \citep{hartmann2025lostinmoderation}. Both lines stop short of a fused multimodal classifier, which is a chain of models: an extraction stage decides what text exists, a cap decides how much survives, two encoders decide what to represent, and a fusion layer decides what to output. A disparity can enter at any link, and an audit reading only the final decision is coarse about \emph{where} it entered and, worse, tests at the link where its own evidence is thinnest.

\textit{This leads to our key question: at which stage of a fused multimodal moderation pipeline can a subgroup asymmetry actually be detected, in what the classifier is permitted to read, in what its branches encode, or in what it finally decides?}
Our axis is source language, so the audit tests fairness across language communities; it makes no claim about demographic groups. We make this testable through three claims, one per stage, each stated over the same 790-clip TikHarm test split. All three are held to one standard. An asymmetry counts only when it clears a significance test; the size of a percentage-point difference alone is not enough.

\emph{First, at the decision stage, disaggregating the 790 logged decisions by source language reveals a flagging-rate gap}, counted real only if it is at least 5 percentage points \emph{and} clears Fisher's exact test at the 5\% level across all pairwise comparisons after Holm correction. \emph{Second, at the representation stage, one pre-fusion branch carries subgroup identity and is the same branch whose removal moves that gap}, credited only if a branch beats \emph{its own} permutation null at the 5\% level \emph{and} removing it collapses the gap while the classifier still discriminates. \emph{Third, at the input stage, the preprocessing chain admits an unequal share of each subgroup's content before the classifier reads any of it}, counted only if the retained-fraction difference clears the 5\% level after Holm correction within its own family. We report each against its criterion, including where it is not met, and read the three together as one pass through the pipeline.

This paper makes three contributions. First, we bring subgroup-disaggregated error-rate auditing to a setting prior work has left open, a fused multimodal moderation classifier, and require the resulting gap to pass a significance test \citep{besse2018confidence, cherian2023statistical, ferrara2025size} before calling it a finding. Second, we pair two modality-attribution techniques, a leave-one-modality-out ablation and a subgroup-identity probe, so a disparity can be localized without either technique's blind spot alone. Third, we run the audit at several stages of the pipeline as well as at its output, extending it to the preprocessing chain that decides what the classifier may read. We release it as a tool that can be pointed at any guard-style system that logs per-item decisions, with no access to that system's internals.\footnote{Code and data: \url{https://github.com/ksubasi/cross-modal-fairness-audit}.}

\section{Background and Related Work}
\label{sec:related}

\textbf{Fairness auditing outside content moderation.} Fairness auditing has repeatedly shown aggregate accuracy concealing large subgroup disparities. Commercial face-analysis systems reached near-perfect accuracy yet erred far more on darker-skinned women than any other subgroup \citep{buolamwini2018gendershades}, and survey work catalogues the stages at which bias enters \citep{mehrabi2021survey}. This literature is largely unimodal, where an error's source is easy to localize to the one model in question. A few studies do disaggregate multimodal systems built for other tasks, finding race the largest disparity axis in mental-health assessment \citep{jiang2023evaluating} and comparable demographic gaps in automated video-interview scoring, where combining modalities fails to reliably improve fairness even when it improves accuracy \citep{booth2021bias}. Both show that subgroup disparities survive fusion; neither is in the moderation domain, and neither localizes a disparity to a stage.

\textbf{Fairness auditing inside content moderation.} The closest work to ours audits five commercial moderation APIs, reporting error rates by target identity group and finding systematic over-moderation of Black, LGBTQIA+, Jewish, and Muslim content alongside under-moderation of others \citep{hartmann2025lostinmoderation}. Those systems are text-only APIs, where a disparity has one place it can come from and one stage at which to test it. Moderation fairness has also been approached through persona-infused and multi-agent framings \citep{gajewska2026fairness, gajewska2026harmful}, which foreground \emph{who} is harmed and leave error-rate disparities in a deployed classifier untouched.

\textbf{Multimodal content moderation without fairness auditing.} Text-only hate-speech detection is a mature area with its own systematic reviews \mbox{\citep{jahan2023systematic}}. Multimodal classifiers fusing video and audio with text are increasingly common and accurate \citep{cespedes2025mmhsd, hee2024recent, lang2025more}, yet they are evaluated purely on aggregate metrics. MTikGuard reports 89.37\% accuracy on extended TikHarm with no subgroup breakdown \citep{nguyen2025mtikguard}, HateMM reports aggregate accuracy and macro-F1 \citep{das2023hatemm}, and MultiHateClip reports a large English-vs-Chinese gap but frames it as model evaluation and draws no fairness conclusion \citep{wang2024multihateclip}. Relatedly, \citet{bui2024multi3hate} finds annotators across cultures disagree on what counts as hateful, making content origin fairness-relevant before a classifier is involved. Each of these evaluations reports aggregate or per-language performance only.

\textbf{Modality attribution and evidence localization.} Methods for localizing a disparity to a modality exist outside moderation. Fairness-weighted fusion down-weights the most disparate modality \citep{hooman2025fame}, and causal fusion intervenes on each modality separately \citep{cheong2024fairrefuse}, both targeting health records and depression detection. Inside hate-video analysis, attribution has been pursued for interpretability \citep{agentic2026evidence}. We adapt the simplest member of this family and make no claim that modality attribution is new; our addition is to hold it to a significance standard alongside a stage it cannot see.

\textbf{Statistical inference for fairness auditing.} A subgroup gap is a point estimate, and the auditing literature warns against treating one as a finding without an interval or a test \citep{besse2018confidence, cherian2023statistical}; small subgroups inflate apparent disparities \citep{ferrara2025size}, and the effective sample size for a rate is the population the rate is computed over, which can be far smaller than the subgroup \citep{singh2023brief}. That last point is the one this paper builds on. It is usually raised as a caveat on a reported number; we treat it instead as a property that varies systematically along a pipeline and therefore determines where an audit should look.

\textbf{Tokenization disparity as a fairness axis.} Multilingual subword tokenizers encode the same content at very different lengths across languages. Across seventeen tokenizers some language pairs differ by as much as fifteen times in tokenized length \citep{petrov2023tokenizers}, and some languages need roughly five times as many tokens to express the same information \citep{ahia2023costsame}. That fragmentation comes from the tokenizer's design, since pre-tokenization choices largely determine whether a script is represented efficiently \citep{velayuthan2025tokenizers}. This literature discusses the consequence as cost, latency, and context-window pressure. We connect it to a moderation decision. When a pipeline caps its text input, a higher-fertility language loses more content before the classifier reads it, making an ordinary engineering constant a fairness-relevant choice.

\textbf{The gap.} Fairness auditing inside moderation has stayed text-only, while multimodal fairness auditing has stayed outside moderation. What neither supplies is a case where one fairness test is run at several stages of a single pipeline and the stages are allowed to disagree, which is what makes it possible to say both whether a disparity is real and where an audit can see it.

\section{The Cross-Modal Fairness Audit}

\begin{figure*}[t]
\centering
\includegraphics[width=\textwidth]{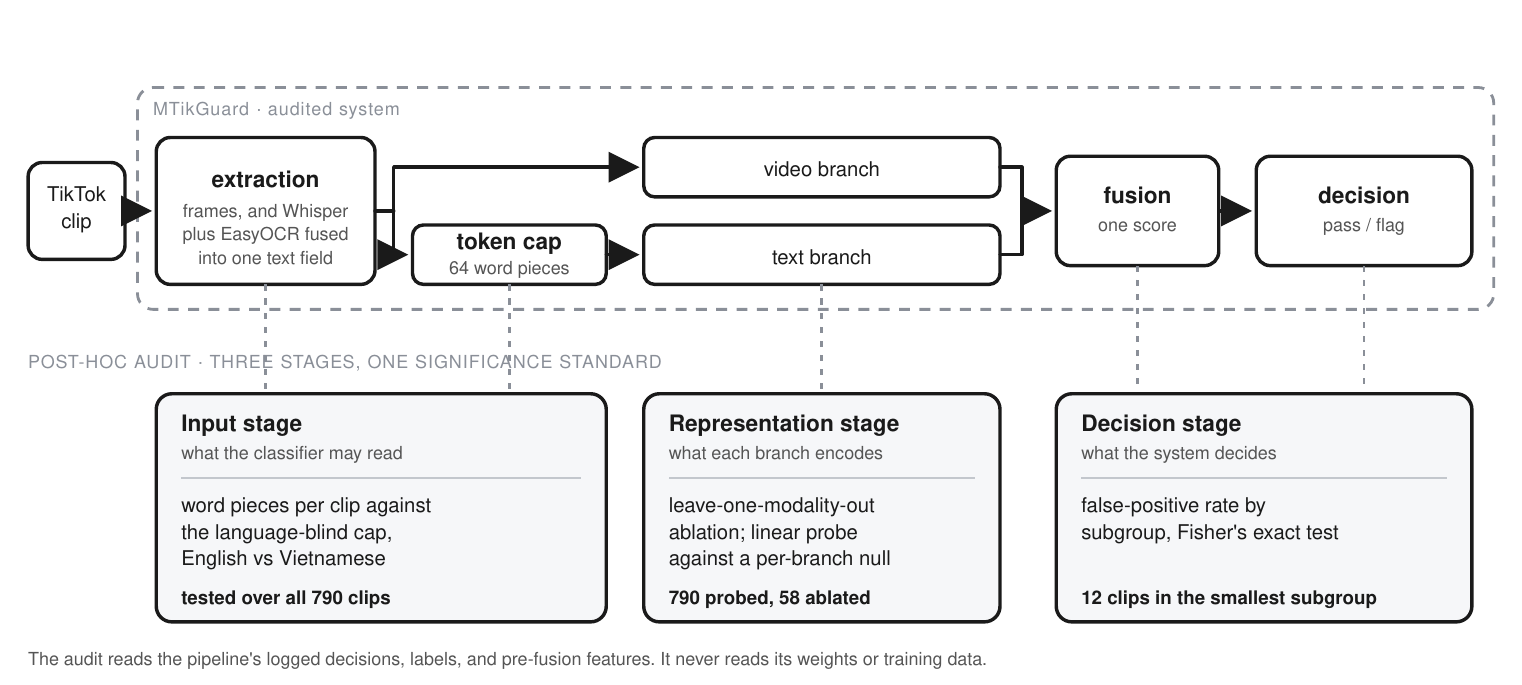}
\caption{The audited pipeline and the three stages at which we test it. A clip enters MTikGuard as a raw video file; extraction yields sampled frames, a Whisper transcript, and an EasyOCR string concatenated into one text field, a language-blind cap decides how much of that field survives, two encoders represent what is left, and a fusion layer emits one label. The audit taps three of these links and holds all three to the same significance standard. The item counts show why they are not equally testable. The cap and the encoders can be tested over every clip, while a false-positive rate exists only over a subgroup's true-\textsc{pass} items, which leaves twelve in the smallest subgroup.}
\label{fig:apparatus}
\end{figure*}

The \emph{cross-modal fairness audit} is a post-hoc procedure for testing whether a deployed multimodal classifier treats language subgroups equally. It runs at three stages of the pipeline, holds each to the same significance standard, and reads only what the system already logs. Figure~\ref{fig:apparatus} shows the audited pipeline, the three stages at which we test it, and the evidence each stage has to work with. Below we describe the framework the audit runs in, the pipeline it is pointed at, the subgroups it splits by, the instruments that localize a disparity, and the metrics that gate one.

\subsection{Auditing Framework}
\label{sec:framework}

The audit takes three inputs: the pipeline's per-item flagging decisions, the labels judging those decisions correct or incorrect, and a subgroup taxonomy. It requires no access to the pipeline's internals, which is deliberate, because platforms are often unwilling to expose weights or training data to an external auditor, so an audit working from decisions and labels alone can be run by a third party without cooperation beyond access to logged outputs.

For each subgroup it computes a false-positive rate over that subgroup's true-\textsc{pass} items and a false-negative rate over its true-\textsc{flag} items, then reports the largest pairwise difference on each metric; the pair achieving that maximum is carried into the significance test. It then repeats the disaggregation at two stages further upstream.

\subsection{Audited Pipeline}
\label{sec:pipeline}

We audit MTikGuard \citep{nguyen2025mtikguard}, a transformer-based multimodal classifier for child-safe TikTok moderation published at PACLIC 39. We audit a published system because others can inspect and rerun it, which makes a fairness claim about it stronger evidence.

It is a fine-tuned encoder pipeline with no prompting, so its behaviour is fixed at training time and its per-branch representations can be read directly. Frames pass through a TimeSformer encoder; Whisper speech and EasyOCR on-screen text are concatenated into one field and read by a multilingual DistilBERT, so there is no separate audio encoder. Late fusion of the two branch encodings feeds a linear layer over four labels, the configuration reporting 89.37\% accuracy on TikHarm.

A clip enters as a raw video file with no uploader, caption, or posting context, so every subgroup signal must come from inside it. The audit reads what the pipeline logs along the way, and these logs are not equally testable. The fused text field exists for every clip, while a false-positive rate exists only over a subgroup's true-\textsc{pass} items. The concatenation costs the audit most, erasing the boundary between speech and on-screen text before the encoder sees it, which is why the ablation below removes a segment of text before tokenization.

Two deviations bear on the results. The preprocessed table MTikGuard trains from was never released, so we rebuild it from the public raw videos following its described extraction logic; what we audit is a faithful reimplementation of the authors' system. For compute reasons we transcribe with Whisper-small and detect language with \texttt{langdetect}, so the subgroup assignment is our own. The token cap follows the original. The paper states no maximum text length, but the released training code fixes one at 64 word pieces with truncation and fixed-length padding, and we adopt that value.\footnote{\texttt{Multimodal Training.ipynb} in the system's public training repository; the cap is the dataset class's \texttt{max\_text\_length}, left at its default in every split.} Section~\ref{sec:trunc} shows that this apparently neutral constant treats the subgroups differently.

\subsection{Subgroup Taxonomy and Data}
\label{sec:taxonomy}

Our two creators differ on one attribute the pipeline is measurably sensitive to, the language its transcription and OCR stages must work in. A taxonomy has to make that explicit, so error rates can be recomputed within each cell.

The corpus is the extended TikHarm dataset released with MTikGuard: TikTok clips labelled with one of the four classes above. The audit runs on the held-out test split of 790 clips, close to balanced across classes (200 Safe, 198 Harmful Content, 197 Suicide, 195 Adult Content).

Our subgroup axis is content origin, specifically source language, already present through TikHarm's bilingual collection and MTikGuard's own language-detection step. The Vietnamese--English contrast comes from the corpus: TikHarm was assembled as a Vietnamese TikTok child-safety collection with English material alongside it, so those are the two languages a published classifier for this corpus was trained on. That detection step computes each clip's language and discards it; we persist it, and that one retained column is what makes the axis available at all. It does not always return a confident label: 74 of 790 test items (9.4\%) carry none and form a third \emph{unknown} subgroup, which we keep and report, since discarding it would itself hide a disparity if the pipeline treats undetected-language content differently. TikHarm carries no native demographic labels, and we did not infer them, since inferred attributes carry well-documented bias risks, and an uncontrolled second axis would trade one caveated result for two.

\subsection{Modality-Attribution Method}
\label{sec:attribution}

To test whether a disparity is concentrated in one input channel we adopt a leave-one-modality-out comparison, re-evaluating the classifier three more times with the trained weights fixed and one modality removed each time. Removing video replaces the sampled frames with a fixed blank frame, matching the pipeline's own null-video fallback. Speech and on-screen text are not encoded separately, since the pipeline fuses them into one string before the text encoder sees it (Section~\ref{sec:pipeline}), so we drop the corresponding segment before tokenization, isolating what each contributes alone. A modality whose removal collapses the gap is implicated as the channel carrying it; one whose removal leaves it unchanged is not. This is the simplest member of the family of Section~\ref{sec:related} \citep{hooman2025fame, cheong2024fairrefuse}, adopted because it requires no change to the trained classifier.

The ablation has a blind spot worth naming before using it. Nulling a modality can erase a subgroup gap simply by erasing the classifier's ability to discriminate at all, collapsing every prediction to one class. To check for this we add a second technique that leaves the classifier intact: a linear probe on each branch's pre-fusion features that predicts subgroup membership, scored by five-fold cross-validation. Its chance baseline is the identical probe on \emph{that same branch's} features with the subgroup labels permuted (50 permutations). Each branch is scored only against its own null, because a branch's chance macro-F1 depends on its own feature geometry, and one pooled baseline would misjudge whichever branch's true chance level differs from it. Probe accuracy is a debated proxy for what a model \emph{uses} a representation for \citep{elazar2021amnesic, voita2020mdl, pimentel2020probing}; our claim is only the coarse one, that subgroup identity is decodable far above a branch's own chance, for which macro-F1 against a per-branch permutation null suffices.

\subsection{Metrics}
\label{sec:metrics}

Since TikHarm's labels are one of four classes, we collapse each prediction and gold label to a binary flagging outcome (\emph{flagged} for Adult Content, Harmful Content, or Suicide, \emph{passed} for Safe), then compute $\mathrm{FPR}_g$ and $\mathrm{FNR}_g$ within each subgroup $g$ alongside the aggregate rates. The quantity we highlight is the maximum pairwise gap $\max_{g,g'} |\mathrm{FPR}_g - \mathrm{FPR}_{g'}|$, since that number says whether the aggregate figure was representative. We do not collapse the subgroup rates back into one combined fairness score, which would reproduce the aggregation problem the audit exists to expose.

A raw percentage-point gap alone is weak evidence that a classifier treats subgroups differently \citep{besse2018confidence, cherian2023statistical}: with several subgroups compared pairwise and a small enough one, some pairing will look large from sampling noise alone \citep{ferrara2025size}. We test the pair achieving the maximum gap with Fisher's exact test, restricted to the population the metric is computed over (true-Safe items for FPR, true-flagged for FNR), since the subgroup's full count overstates the effective sample size \citep{singh2023brief}. Reporting only the arg-max pair would leave that test selected for its extremity, so we run Fisher's exact on \emph{every} pairwise comparison and adjust the family by Holm--Bonferroni. The 5-point materiality floor guards against dignifying a trivial difference with a test; it never binds here, since the largest gap clears it comfortably and fails on significance instead.

\section{Experimental Setup}

We evaluate on the held-out split released with MTikGuard: 790 videos, unchanged from the original TikHarm test set so our aggregate accuracy is directly comparable to the reported one. Its language-detection step yields 482 Vietnamese, 234 English, and 74 undetected items; within the 200 true-Safe items the false-positive comparison runs over, those subgroups hold 142, 46, and 12 respectively.

We fine-tune the audited classifier on the 2,762-video training split under the published hyperparameters (learning rate $1\times10^{-4}$, effective batch size 16, 6 epochs, checkpoint by highest validation macro-F1), then run the audit and the three leave-one-modality-out variants on the test split. Because we reproduce the published configuration, any gap belongs to the system under study and cannot be blamed on our training choices.

\section{Results}

We report the three stages in the order an auditor meets them, starting where a conventional subgroup audit stops, at the model's logged decisions, and then moving back up the pipeline to the representation and the input. Each stage is judged against the criterion set for it in the introduction, and the three are then read together.

\subsection{The Decision Stage}
\label{sec:decision}

Table~\ref{tab:subgroup} reports error rates for the three source-language subgroups. The unknown-language subgroup has the highest false-positive rate, 7.97 points above English (the pair anchoring the full-model FPR gap); the false-negative gap runs the other way, Vietnamese highest at 4.71 points above unknown. Subgroup accuracy stays within a 3.52-point band (93.78\%--97.30\%), well under either gap, so no subgroup is simply harder to classify overall.

\begin{table}[t]
\centering
\small
\begin{tabular}{lcccc}
\toprule
\textbf{Subgroup} & \textbf{FPR} & \textbf{95\% CI} & \textbf{FNR} & \textbf{N} \\
\midrule
Vietnamese & 9.86\% & [6.0, 15.9] & 4.71\% & 482 \\
English & 8.70\% & [3.4, 20.3] & 2.13\% & 234 \\
Unknown & 16.67\% & [4.7, 44.8] & 0.00\% & 74 \\
\midrule
Aggregate & 10.00\% & [6.6, 14.9] & 3.39\% & 790 \\
\bottomrule
\end{tabular}
\caption{Disaggregated error rates by source-language subgroup on the 790-item held-out test split. Unknown = items with no confidently detected language (Section~\ref{sec:taxonomy}). The 95\% Wilson intervals are computed over each subgroup's true-\textsc{pass} items, which number 142, 46 and 12 respectively; the unknown subgroup's interval therefore spans 5--45\%. All three overlap, which is why the 7.97-point point-estimate gap does not survive Fisher's exact test ($p=0.59$; Vietnamese-vs-unknown FNR $p=0.15$; all six pairwise tests Holm-adjusted $p \ge 0.44$).}
\label{tab:subgroup}
\end{table}

Neither gap clears the significance check. The unknown-vs-English FPR comparison rests on only 12 true-Safe unknown items (2 false positives) against 46 true-Safe English items (4 false positives), and the Vietnamese-vs-unknown FNR comparison also falls short (Table~\ref{tab:subgroup}). Testing every pairwise comparison changes nothing: none of the six reaches significance on either metric, before or after Holm correction. With a subgroup this small, one item moves its rate by several points, and the test split, fixed by the original TikHarm release, cannot be enlarged to settle it. A simulation puts the power of the observed comparison at about 11\%. With 12 and 46 true-Safe items, Fisher's exact test reaches 80\% power only if the unknown subgroup's FPR is around 45--50\%, a gap of roughly 36--41 points, and the Vietnamese--English comparison only at a Vietnamese FPR of about 27--30\%; Holm correction makes both stricter. Against its stated criterion, \textbf{the decision-stage gap does not count as real}: the 7.97-point gap clears the 5-point size bar but not the significance one.

The 58 items carrying that comparison are worth holding on to, because they are 7\% of the 790 the pipeline processed.

\subsection{The Representation Stage}
\label{sec:representation}

Table~\ref{tab:attribution} recomputes the gap under each leave-one-modality-out configuration. Removing either half of the text channel leaves the gap unchanged (no subgroup rate moves by more than about two points), so neither transcribed speech nor on-screen text is where the disparity originates. Removing video is the only intervention that changes the gap, and it erases it completely, but only by collapsing the classifier itself. Accuracy falls from 94.94\% to 74.68\% (the flagged-class base rate) and every item is predicted flagged. A classifier with no signal left cannot produce a subgroup gap at all, so the zero shows that the pipeline has no usable signal without video, and says nothing about whether video is fair. The ablation therefore narrows localization without settling it.

\begin{table}[t]
\centering
\small
\setlength{\tabcolsep}{4pt}
\begin{tabular}{lccc}
\toprule
\multicolumn{4}{l}{\emph{Ablation}: gap with a modality nulled} \\
\midrule
 & FPR gap & FNR gap & Accuracy \\
Full model & 7.97 & 4.71 & 94.94\% \\
$-$ Video & 0.00 & 0.00 & 74.68\% \\
\midrule
\multicolumn{4}{l}{\emph{Probe}: subgroup identity per branch} \\
\midrule
 & macro-F1 & own null & $p$ \\
Text branch & 0.71\,$\pm$\,0.04 & 0.30\,$\pm$\,0.014 & $\leq$0.02 \\
Video branch & 0.34\,$\pm$\,0.03 & 0.33\,$\pm$\,0.02 & 0.33 \\
\bottomrule
\end{tabular}
\caption{The two attribution instruments at the representation stage. Gaps are percentage points. Nulling audio or on-screen text leaves both gaps unchanged and is omitted. Removing video zeroes the gap only by collapsing the classifier to always-flag, which is what the accuracy column shows. Each probe branch is scored against \emph{its own} permuted-label null, since a branch's chance level depends on its own feature geometry; $\leq$0.02 is the floor attainable with 50 permutations.}
\label{tab:attribution}
\end{table}

The probe runs over all 790 items, against the 58 the ablation's gap rests on, and it gives a clear answer. Recovering subgroup membership from each branch's pre-fusion representation, it reaches macro-F1 $0.71 \pm 0.04$ on the text branch against that branch's own permuted baseline of $0.30 \pm 0.014$, roughly 29 standard deviations above it. The video branch reaches $0.34 \pm 0.03$ against its own baseline of $0.33 \pm 0.02$, half a standard deviation above it, which is indistinguishable from chance. Subgroup identity is thus strongly encoded in text, with no detectable trace in video.

Against its stated criterion the representation-stage claim \textbf{splits}: a branch does carry subgroup identity and clears its own null decisively, while the branch whose removal moves the gap is a different one, and its removal moves the gap only by destroying the classifier. The joint claim fails while each half stands on its own evidence. The two instruments answer over populations of very different size, and the one with 790 items behind it returns a signal.

\subsection{The Input Stage}
\label{sec:trunc}

Both stages above hold the model responsible for any disparity. One step before the model, the fused text field is truncated at 64 word pieces, and while that cap is language-blind, word-piece segmentation depends on the language. This stage is tested over the extracted text of all 790 clips, the largest population available anywhere in the audit.

\begin{figure}[t]
\centering
\includegraphics[width=\columnwidth]{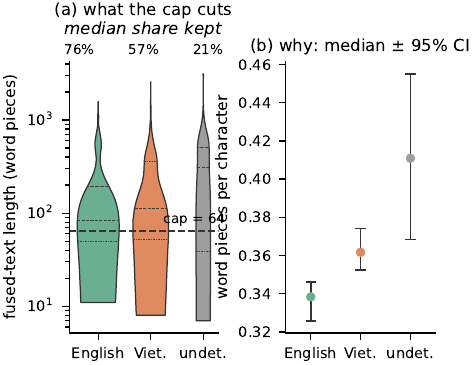}
\caption{What a language-blind token cap does to each subgroup. (a) Fused-text length per clip (log scale) against the pipeline's 64-word-piece cap, where everything above the line is discarded before the classifier reads it, with the median share each subgroup keeps printed above. (b) The cause is segmentation: median word pieces per character with bootstrap 95\% intervals, Vietnamese separated from English while the volume of text extracted does not differ. Tests (Holm-adjusted within the four-test family): truncated at all, Fisher $p=0.21$; retained fraction, Mann--Whitney $p=0.014$, $p_{\text{adj}}=0.042$; word pieces per character, $p=3\times10^{-4}$, $p_{\text{adj}}=0.001$; extracted volume, $p=0.18$. The undetected-language subgroup is the most fragmented and most truncated, and its wide interval is the same small-subgroup problem that defeats the comparison in Section~\ref{sec:decision}.}
\label{fig:truncation}
\end{figure}

Whether a clip is truncated at all barely differs by subgroup: 66.8\% of Vietnamese clips exceed the cap against 62.0\% of English ones, a 4.8-point gap that fails the same check the FPR gap failed. How much survives the cap does differ (Figure~\ref{fig:truncation}): a Vietnamese clip retains a median 57.4\% of its extracted text against 76.2\% for an English clip. The cause is segmentation: Vietnamese costs 6.9\% more word pieces per character, while the volume of text extracted is the same. These four comparisons form one family, so we correct them together as above. After Holm adjustment the segmentation-rate result stands comfortably and the retained-fraction result stands narrowly, while the other two fall short (Figure~\ref{fig:truncation}). Against its stated criterion, \textbf{the input-stage asymmetry counts as real}.

It is a disparity in what the classifier is permitted to read; we have not shown a disparity in what it decides. We did not retrain under a larger cap, so we cannot say it moves any flagging outcome.

\section{Discussion}

The traversal returns a graded answer, and the grade tracks the evidence available at each stage. At the input, over 790 clips, a cross-language asymmetry clears its significance check; at the representation, over the same 790 items, subgroup identity is decodable from the text branch at roughly 29 standard deviations above its own null. An audit reading only decisions would have found neither. The two also differ in what they cost to fix: an asymmetry at the input is an engineering constant someone can change in an afternoon, one in a branch is a retraining project, and one visible only at the decision is a measurement problem before it is a fairness problem. Because the audit reads only logged outputs, a trust-and-safety team can run it in place, and an external researcher wherever a platform releases those logs.

This audit leaves three questions open. The decision-stage gap of 7.97 points is indistinguishable from noise over 58 items, so our two creators' divergent fates cannot be confirmed where the moderation decision is made. The Vietnamese creator's clip nonetheless demonstrably reaches the classifier with less of its speech and on-screen text intact. The two attribution instruments disagreed, leaving neither branch cleanly responsible nor cleanly innocent. And this data cannot separate two readings of the whole pattern: the asymmetry may attenuate downstream, with a model steady enough that unequal input does not produce unequal output, or it may persist and simply be undetectable where the test has 58 items. A null at the decision stage is usually written up as evidence a classifier is fair. This one sits alongside a significant asymmetry two stages upstream, on the same axis and over the same clips, and that combination can only be interpreted if an audit reports the population each test ran over \citep{singh2023brief, ferrara2025size}.

The most direct next step is settling that ambiguity by retraining under a larger cap, which would show whether the input-stage asymmetry propagates to a flagging outcome. Beyond it, the taxonomy extends naturally to the target-group and platform-of-origin axes in datasets such as DeHate \citep{zhang2025dehate}, whose per-video annotations of which modality carries the harmful content would supply the ground truth our attribution instruments had to work without. The same staging applies beyond automated moderation, for instance to multi-agent systems that train teachers to manage hate incidents \citep{gajewska2025teachers} or tutor students through ethical debate \citep{maslowski2026debate}, where it would show whether what such a system treats as harmful depends on the language the material arrives in.

\section{Conclusion}

Multimodal classifiers now decide what may be seen on short-video platforms, and one aggregate accuracy figure is compatible with error falling almost entirely on a single language group. We asked at which stage of a fused pipeline a subgroup asymmetry can be detected at all.

Auditing MTikGuard, a published TikTok classifier, we held all three stages to one significance standard. At the decision the observed false-positive-rate gap did not survive, resting as it did on 58 items. At the representation a probe found subgroup identity strongly encoded in the text branch and, against the correct per-branch baseline, at chance in the video. Removing video collapsed the classifier to always-flag, so neither branch came out cleanly responsible or innocent. At the input the language-blind cap discarded a significantly larger share of a Vietnamese clip than an English one. The verdicts ordered themselves exactly as the sample sizes did.

Power falls along the pipeline together with stage position, so we cannot rule out a genuine decision-stage disparity that is merely unmeasurable in a split this small. The instruments have complementary blind spots: the ablation is coarse, and the probe shows what a branch encodes, which may differ from what the classifier uses. The dataset's lack of native demographic labels also leaves source language as our only gold-standard axis. The audit covers one pipeline, so this demonstrates the method; it does not establish a general property of multimodal moderation.

The contribution is a discipline more than a datum. Where an audit tests along a pipeline determines how much evidence it has, and the logged decisions most subgroup audits read are where that evidence is thinnest. Staging the same audit upstream cost no extra access and recovered an asymmetry the decision-level view returned as noise. As moderation decides which voices reach an audience, telling real harm from a sampling artefact, and naming the stage it came from, is what holds these systems to account.

\section*{Acknowledgments}
This work was partly supported by the VolkswagenStiftung under grant No.~98542 (project Deliberation Laboratory, DeLab).

\bibliography{references}

\end{document}